\documentclass{article}

\usepackage{arxiv}
\usepackage{amsmath} 
\usepackage[utf8]{inputenc} 
\usepackage[T1]{fontenc}    
\usepackage{hyperref}       
\usepackage{url}            
\usepackage{booktabs}       
\usepackage{amsfonts}       
\usepackage{nicefrac}       
\usepackage{microtype}      
\usepackage{lipsum}		
\usepackage{graphicx}
\usepackage{natbib}
\usepackage{doi}

\title{3D2M Dataset: A 3-Dimension diverse Mesh Dataset}

\author{ \href{https://orcid.org/0000-0000-0000-0000}{\includegraphics[scale=0.06]{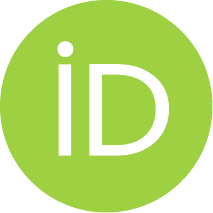}\hspace{1mm}Sankarshan Dasgupta}\\
	Department of Computer Science\\
	University of Dayton\\
	Dayton, OH 45469 \\
	\texttt{dasguptas2@udayton.edu} \\
}

\renewcommand{\shorttitle}{\textit{arXiv} Template}

\hypersetup{
pdftitle={3DM Dataset: A diverse 3-Dimension Mesh Dataset},
pdfsubject={q-bio.NC, q-bio.QM},
pdfauthor={Sankarshan Dasgupta},
pdfkeywords={3D mesh, .obj file, facial keypoints location, Mediapipe},
}

\begin{document}
\maketitle

\begin{abstract}
Three-dimensional (3D) reconstruction has emerged as a prominent area of research, attracting significant attention from academia and industry alike. Among the various applications of 3D reconstruction, facial reconstruction poses some of the most formidable challenges. Additionally, each individual’s facial structure is unique, requiring algorithms to be robust enough to handle this variability while maintaining fidelity to the original features. This article presents a comprehensive dataset of 3D meshes featuring a diverse range of facial structures and corresponding facial landmarks. The dataset comprises 188 3D facial meshes, including 73 from female candidates and 114 from male candidates. It encompasses a broad representation of ethnic backgrounds, with contributions from 45 different ethnicities, ensuring a rich diversity in facial characteristics. Each facial mesh is accompanied by keypoints that accurately annotate the relevant features, facilitating precise analysis and manipulation. This dataset is particularly valuable for applications such as facial re targeting, the study of facial structure components, and real-time person representation in video streams. By providing a robust resource for researchers and developers, it aims to advance the field of 3D facial reconstruction and related technologies.

\end{abstract}


\section{Introduction}
Three-dimensional reconstruction has become a focal point of research in computer vision, with applications spanning virtual reality, augmented reality, animation, and more. Among the various objects and structures that can be reconstructed, human faces present unique challenge due to their intricate geometries and the need for high fidelity in representing subtle features as per \cite{Faceperception} et al. Accurate facial reconstruction is essential for numerous applications, \cite{LIANG202012} et al. and \cite{MORALES2021100400} gives prior examples of facial reconstruction requirements in diverse applications including identity verification in [\cite{hong2022facial}, \cite{winter1994verification}], emotion recognition in \cite{chen2020cnn}, and realistic character animation in [\cite{pan2023real}, \cite{thambiraja2023imitator}] and other applications. By providing this dataset, we aim to facilitate advancements in facial reconstruction techniques and support the development of more inclusive and accurate models in the field.

The diversity and comprehensiveness of the dataset are its main advantages. It guarantees a comprehensive portrayal of human facial anatomy with its 188 3D facial meshes, which reflect different ethnic backgrounds and gender identities. \cite{masi2016we} et al. have suggested diversity enables researchers to train less biased and more inclusive models, which is essential for creating reliable algorithms for facial recognition and reconstruction . Additionally, \cite{khabarlak2021fast} states the inclusion of facial landmarks enhances the dataset's utility by providing precise annotations for key features, making it easier to study facial dynamics. This combination of diversity, encompassing individuals aged 16 to 65, and detailed annotation makes the dataset an invaluable resource for advancing research in areas such as real-time facial representation. We have studied the complex nature of human facial structures, which is essential for creating accurate and reliable models in 3D reconstruction. By providing detailed annotations, including key facial landmarks, we empower researchers to delve deeper into the intricacies of facial dynamics across different age groups. This inclusive approach enhances the development of effective facial recognition technologies that are adaptable to a broader demographic spectrum.

\begin{figure}[h]
  \centering
  \includegraphics[width=\linewidth]{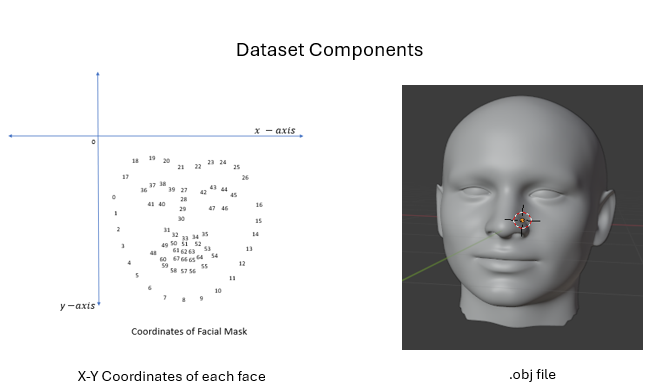}
  \caption{The dataset consist of .obj file and keypoints vector file}
   \label{fig:dataset}
\end{figure}

We have studied the requirements for 3D reconstructions, emphasizing the importance of keypoints and accurate mesh models. As illustrated in Figure \ref{fig:dataset}, the mesh and keypoints coordinates dataset with diverse ethnic features can be extremely beneficial for real-time reconstructions in the near future.

\section{Related Works}
In recent years, 3D reconstruction has thrived \cite{flores20233d}, leveraging various methodologies that utilize 3D Morphable Models (3DMM) is detailed in \cite{ning2023multi}, Neural Radiance Fields (NeRF) in \cite{zhang2022fdnerf}, Generative and deep CNN algorithms [\cite{toshpulatov2021generative}, \cite{zhang2021brief}], and Image processing based algorithm in \cite{SanDasgupta}. These approaches have demonstrated remarkable effectiveness in capturing intricate details and spatial relationships in 3D models, enabling more accurate and efficient reconstructions. Although these models are becoming increasingly powerful, the training process still requires a diverse and comprehensive dataset \cite{li2022comprehensive} has mentioned to effectively capture the complexities of 3D geometry.

Several prominent face reconstruction datasets, including AFLW2000-3D \cite{Zhu_2016_CVPR}, FaceWarehouse \cite{6654137}, the Florence3D Faces dataset \cite{6595917}, and VGGHeads \cite{kupyn2024vggheadslargescalesyntheticdataset}, have significantly contributed to approaches in this field. AFLW2000-3D contains 2,000 images annotated with 68-point 3D facial landmarks, providing a robust resource for studying facial structures. FaceWarehouse offers a comprehensive database of 3D facial geometries for 150 subjects, focusing on various facial expressions. The Florence 3D Faces dataset pairs 3D models with corresponding 2D videos, ensuring consistency across data types. VGGHeads is a large-scale synthetic dataset designed specifically for 3D human heads, providing extensive variations in head poses and expressions, which is invaluable for training deep learning models. Other noteworthy datasets include the DAD-3DHeads dataset \cite{martyniuk2022dad3dheadslargescaledenseaccurate}, which comprises 44,898 images collected from diverse sources for training, validation, and testing purposes. Additionally, FaMoS, \cite{bolkart2023instantmultiviewheadcapture} features dynamic 3D head data from 95 subjects, each performing 28 motion sequences. Lastly, the HEADSET dataset \cite{Lohesara_2023} provides volumetric representations of human interactions, emphasizing human emotion awareness in scenarios involving partial occlusions.

The reconstruction of facial features necessitates high-quality input from a dataset, complemented by advanced machine learning techniques \cite{fahim2021single}. While methods such as Neural Radiance Fields (NeRF) can operate without a pre-defined 3D face mesh \cite{feng20243d}, generative networks and Convolutional Neural Networks (CNNs) rely on the diverse features contained within a comprehensive 3D dataset \cite{yang2020facescape}. Previous works suggests, datasets may offer trainable examples, they often fail to include adequate representation of multiethnic faces, thereby limiting their applicability in real-world scenarios \cite{li2022comprehensive}. By simplifying usability and enhancing the understanding and recovering 3D meshes \cite{10195242} and keypoints in \cite{10.5555/1577069.1755843}, we aim to empower researchers and practitioners to advance their work in facial reconstruction and related applications.This effort serves as a vital resource for training models that generalize better across diverse populations, ultimately contributing to the development of more inclusive and accurate facial recognition technologies \cite{leslie2020understanding}. This gap motivated us to develop a new dataset designed to meet the training requirements for 3D reconstruction while enhancing the usability and understanding of 3D meshes and keypoints.

Thus, we conclude that our dataset has the potential to significantly enhance 3D mesh reconstruction in \cite{KeenToolsFaceBuilder} and the accuracy of facial keypoints for neural network-based algorithms. This dataset can lead to more precise training outcomes in 3D reconstruction techniques as mentioned for quality assesment in \cite{10.1145/3592786}, effectively addressing the challenges present in available 3D datasets. This dataset has been utilized in my previously published work on 3D reconstruction.

\section{Methodology}

3D face reconstruction can be achieved through various methodologies, with generative networks being among the most common approaches \cite{kammoun2022generative}. These networks leverage extensive datasets to enhance their performance and accuracy. Our face mesh model dataset features a diverse array of facial structures representing 45 different ethnic backgrounds. It is organized into 118 distinct classes, comprising 73 models from female candidates and 114 from male candidates. This careful categorization not only enriches the dataset but also ensures comprehensive representation, 
making it a valuable resource for advancing 3D face reconstruction techniques. The dataset class can be subdivided into:
\begin{itemize}
\item {Dataset Classification} \item {3D Mesh Models}  \item {Feature Selection}
\end{itemize}

\subsection{Dataset Classification}
The dataset comprises a total of 188 3D facial meshes. This collection includes a diverse representation of candidates, with 73 meshes from female candidates and 114 meshes from male candidates. The dataset aims to capture a broad spectrum of ethnic backgrounds, featuring contributions from 45 different ethnicities.

\begin{table}[ht]
  \centering 
  \caption{Folder Structure of the 3D Facial Mesh Dataset}
  \label{tab:folder_structure}
  \begin{tabular}{|l|l|} 
    \hline
    \textbf{Dataset} & \textbf{Folder Structure} \\ 
    \hline
    3D Facial Meshes & MaleFaces: \{MaleFace1, MaleFace2, \ldots, MaleFace114\} \\ 
    & FemaleFaces: \{FemaleFace1, FemaleFace2, \ldots, FemaleFace73\} \\ 
    \hline
  \end{tabular}
\end{table}

\begin{table}[ht]
  \caption{Ethnicity by Region in Dataset}
  \label{tab:ethnic_groups}
  \resizebox{\textwidth}{!}{%
  \begin{tabular}{|l|l|}
    \toprule
    \textbf{Region} & \textbf{Ethnicity} \\ 
    \midrule
    Northern Europe & Scandinavian, Finnish, Icelandic, Sami, Estonian \\ 
    Western Europe  & German, Dutch, French, Irish, Scottish, Welsh \\ 
    Southern Europe & Italian, Spanish, Portuguese, Greek, Catalan, Basque, Sicilian \\ 
    Eastern Europe  & Polish, Ukrainian, Russian, Czech, Slovak, Hungarian, Bulgarian, Romanian, Serbian, Croatian \\ 
    Central Europe  & Austrian, Swiss, Slovenian, Bosnian \\ 
    Africa          & Zulu, Yoruba, Berber, Amhara, Somali \\ 
    Asia            & Chinese, Tamil, Punjabi, Kazakh, Khmer, Kashmiri, Gondi, Malay, Malayali, Manipuri, Sinhalese, Bhutia \\ 
    Latin America   & Mestizo, Afro-Latino, Quechua \\ 
    Middle East     & Kurdish, Assyrian, Druze \\ 
    \bottomrule
  \end{tabular}%
  }
\end{table}

From Table \ref{tab:ethnic_groups}, we can observe diversity in the meshes face structure. This is crucial for applications in facial recognition, computer graphics, and other fields where varied facial characteristics can enhance model robustness and performance. By including a wide range of ethnicity, the dataset ensures that the resulting models are more representative and less biased, promoting fairness in technology deployment. The dataset was meticulously assembled from a variety of sources, featuring contributions from individuals participating in the mesh generation process, alongside images sourced from Google.

\subsection{3D Mesh Models}
The mesh is a OBJ file format which is simplified format as mentioned by \cite{liu20213d}. This file contains approximately 17,000 vertex points connected with faces, each defined with vertex normals specified by the vn directive in the OBJ file format. Each normal vector, with three components (x, y, z) ranging from -1 to 1, plays a crucial role in determining how light interacts with the surface, affecting shading and rendering. This detailed geometry is constructed using FaceBuilderHead, a specialized tool for creating realistic human head models, which results in a complex topology that captures intricate facial features. Each OBJ file is around 3 MB in size, reflecting the high level of detail and the extensive information contained within, including vertex positions and normals. Overall, the combination of a high vertex count and well-defined normals is essential for achieving smooth shading and realistic lighting in 3D graphics.

In Figure \ref{fig:dataset}, mesh model serves as an example of the facial structure included in the dataset. This model showcases the intricacies of human facial anatomy, reflecting the detailed topology and geometry captured through the FaceBuilderHead tool. Its realistic representation highlights key features such as the contours of the cheeks, the shape of the nose, and the structure of the jaw, making it a valuable resource for applications.

\subsection{Feature Selection}

Feature selection is performed using the dlib 68-landmark model, which identifies key facial landmarks and provides their 2D coordinate points. These coordinates are meticulously illustrated and stored within the folder structure alongside the corresponding OBJ files. As demonstrated in Figure \ref{fig:dataset}, the point-based indexing system highlights the precise locations of each landmark on the face, allowing for clear visualization and analysis of facial features. This systematic approach enhances the understanding of facial geometry and facilitates various applications in fields such as computer vision, animation, and facial recognition.

\section{Conclusion and Future Works}
In conclusion, this dataset represents a comprehensive collection of 3D facial models constructed through meticulous mesh modeling techniques, featuring nearly 17,000 vertex points and corresponding vertex normals for enhanced surface detail. The integration of dlib's 68-landmark model for feature selection allows for precise identification and visualization of key facial landmarks, facilitating a deeper understanding of facial geometry. For further exploration, the dataset can be accessed at \url{https://github.com/sohomd/3D2M-Dataset.git}.

we plan to enhance this dataset by incorporating texture maps for each face model. By applying realistic textures, we aim to enrich the visual fidelity and realism of the 3D representations, making them suitable for a broader range of applications, including computer graphics, virtual reality, and facial recognition systems. This development will not only improve the aesthetic quality of the models but also provide valuable data for research and practical implementations in various fields. Additionally, the addition of textures will elevate the dataset's utility, enabling more sophisticated analyses and applications in face-related technologies.

\bibliographystyle{unsrtnat}
\bibliography{references}  






\end{document}